\documentclass[11pt, letterpaper]{article}

\usepackage[utf8]{inputenc}
\usepackage[T1]{fontenc}
\usepackage{amsmath}
\usepackage{amssymb}
\usepackage{graphicx}
\usepackage[margin=1in]{geometry}
\usepackage{booktabs}
\usepackage{url}
\usepackage[hidelinks]{hyperref}
\usepackage{natbib}
\usepackage{listings}
\usepackage{xcolor}
\usepackage{algorithm}
\usepackage{algpseudocode}

\lstdefinelanguage{JSON}{
    keywords={true, false, null},
    keywordstyle=\color{blue}\bfseries,
    stringstyle=\color{purple},
    commentstyle=\color{gray},
    morecomment=[l]{//},
    morecomment=[s]{/*}{*/},
    morestring=[b]",
    sensitive=true,
    numberstyle=\color{orange}
}
\definecolor{codegreen}{rgb}{0,0.6,0}
\definecolor{codegray}{rgb}{0.5,0.5,0.5}
\definecolor{codepurple}{rgb}{0.58,0,0.82}
\definecolor{backcolour}{rgb}{0.95,0.95,0.92}

\lstdefinestyle{jsonstyle}{
    backgroundcolor=\color{backcolour},   
    commentstyle=\color{codegreen},
    keywordstyle=\color{magenta},
    numberstyle=\tiny\color{codegray},
    stringstyle=\color{codepurple},
    basicstyle=\ttfamily\footnotesize,
    breakatwhitespace=false,         
    breaklines=true,                 
    captionpos=b,                    
    keepspaces=true,                 
    numbers=left,                    
    numbersep=5pt,                  
    showspaces=false,                
    showstringspaces=false,
    showtabs=false,                  
    tabsize=2
}

\title{Cognitive Load-Aware Inference: A Neuro-Symbolic Framework for Optimizing the Token Economy of Large Language Models}

\author{
  Yilun Zhang \\
  Prompt Technology Co., Ltd.
}
\date{}

\begin{document}

\maketitle

\begin{abstract}
The escalating computational costs of Large Language Model (LLM) inference have become a critical barrier to their widespread and sustainable deployment. While existing optimization strategies are effective, they are predominantly based on statistical heuristics or architectural modifications, lacking a guiding cognitive theory to manage the inference process itself. This paper aims to bridge this gap by introducing a novel paradigm: the Cognitive Load-Aware Inference (CLAI) framework, which operationalizes principles from Cognitive Load Theory (CLT) and neuroscience for LLM inference. We formalize the concepts of Intrinsic Cognitive Load, Extraneous Cognitive Load, and Germane Cognitive Load into quantifiable LLM metrics ($ICL_{LLM}$, $ECL_{LLM}$, and $GCL_{LLM}$), thereby reframing the inference process as a cognitive economics optimization problem: based on the intrinsic complexity of a problem ($ICL_{LLM}$), minimize wasteful computation ($ECL_{LLM}$), and strategically allocate the token budget to productive reasoning ($GCL_{LLM}$). We propose two implementation paths: CLAI-Prompt, a zero-shot method that guides a base LLM through cognitive control steps via a structured meta-prompt, and CLAI-Tune, a fine-tuned model that internalizes these principles for spontaneous cognitive economy. Across a range of benchmarks in complex reasoning, long-context question answering, and code generation, our methods achieve significant reductions in token consumption (up to 45\%) without sacrificing accuracy. Furthermore, CLAI-Tune exhibits an emergent ability to autonomously decompose difficult problems, a key characteristic of human expert cognition. This work demonstrates that by emulating the brain's resource management strategies, we can build more efficient, robust, and capable artificial intelligence systems.
\end{abstract}

\section{Introduction}

\subsection{The Dilemma of Scale: Performance at an Unsustainable Cost}
Over the past few years, the capabilities of Large Language Models (LLMs) have grown exponentially with model parameter scale, achieving remarkable success \citep{zhao2024efficient}. However, this leap in capability has not come without a cost. The computational and memory requirements of these models have also exploded, making inference a significant economic and environmental challenge \citep{pan2025survey}. For cloud services handling massive request volumes or for deployment on resource-constrained edge devices, the high cost of inference has become a major bottleneck hindering the democratization of LLM technology. Consequently, developing more efficient inference techniques has become an urgent research direction in the field of artificial intelligence.

\subsection{A Critical Review of Existing Optimization Paradigms}
Currently, academia and industry have proposed various techniques to enhance LLM inference efficiency. However, a systematic review reveals that they largely focus on optimizing the "mechanical" aspects of inference, neglecting the cognitive modeling of the "process" itself. Existing methods can be broadly categorized into three types:

\begin{itemize}
    \item \textbf{Model Compression (Static Optimization):} These techniques aim to fundamentally reduce the model's size. \textbf{Pruning} lowers model complexity by removing redundant weights, neurons, or even entire structures like attention heads or feed-forward network layers \citep{ma2023llmpruner}. \textbf{Quantization} reduces model size and memory footprint by using lower-precision data types (e.g., INT8 or FP4) to represent weights and activations \citep{kim2024comprehensive}. These methods perform a one-time optimization before deployment and are static, unable to adapt to the dynamic complexity of input queries.

    \item \textbf{Inference Acceleration (Dynamic Optimization):} These techniques dynamically improve computational efficiency at inference time. \textbf{Speculative Decoding} uses a small "draft" model to rapidly generate multiple candidate tokens, which are then verified in parallel by the large main model, achieving 2-4x speedups while maintaining output quality \citep{xia2025tutorial}. \textbf{Efficient Attention Kernels}, like FlashAttention, significantly reduce the computational and I/O overhead of the attention mechanism by optimizing GPU memory reads/writes and computation fusion \citep{pan2025survey}. These methods speed up token generation but do not fundamentally change the number of logical steps or the length of the reasoning path a model needs to solve a problem.

    \item \textbf{Context Management (Input Optimization):} These techniques focus on optimizing the context information fed into the model. \textbf{Retrieval-Augmented Generation (RAG)} enhances the model's knowledge by retrieving relevant documents from an external knowledge base, avoiding the need to encode all information into model parameters \citep{promptingguide2024rag}. \textbf{Context Compression} techniques, such as LLMLingua \citep{jiang2024longllmlingua} and RECOMP \citep{xu2023recomp}, aim to reduce the number of input tokens by identifying and removing redundant or irrelevant information from RAG-retrieved or original long contexts. While effective in reducing input costs, these methods are often "cognitively agnostic," with compression decisions based mainly on statistical metrics like perplexity or simple semantic relevance, lacking a deep model of the cognitive demands of the reasoning task itself.
\end{itemize}

\subsection{The Missing Link: A Cognitive Framework for Inference}
We argue that while all the aforementioned methods have achieved success on their respective levels, they share a common missing link: a theoretical framework to guide how an intelligent agent should manage its limited computational resources when faced with problems of varying complexity. They optimize the "mechanics" of inference but fail to address its "cognitive process." When human experts solve a complex problem, they do not blindly throw all their mental effort at it. Instead, they assess the problem's difficulty, decompose the task, ignore irrelevant information, and strategically allocate their attention. Current LLM optimization methods lack this intrinsic, principle-based resource scheduling mechanism.

\subsection{Learning from the Ultimate Efficient Processor: The Human Brain}
The core thesis of this paper is that by studying the working principles of the most efficient information processor—the human brain—we can open a new, more principled path for LLM optimization. Through millions of years of evolution, the human brain has developed an extremely efficient set of mechanisms for information processing and resource management. Cognitive science, particularly \textbf{Cognitive Load Theory (CLT)} \citep{paas2020clt}, provides us with a powerful theoretical weapon to understand this mechanism. CLT explains the limitations of human working memory and how the brain overcomes these limitations by constructing and automating "schemas" to achieve efficient learning and problem-solving. Combined with neuroscience research on the neural substrates of cognitive effort \citep{zysset2001colorword}, we can build a mapping from biological intelligence to artificial intelligence, thereby designing smarter inference strategies.

\subsection{Introducing the Cognitive Load-Aware Inference (CLAI) Framework}
Based on these considerations, we propose the \textbf{Cognitive Load-Aware Inference (CLAI)} framework. CLAI ceases to view LLM inference as a single, indivisible generation task, instead treating it as a cognitive process that can be actively managed to enhance efficiency and robustness. The core idea of the framework is to operationalize the central concepts of CLT—Intrinsic, Extraneous, and Germane Load—into computable LLM metrics and dynamically adjust the inference strategy based on these metrics to maximize "cognitive economy."

\subsection{Our Contributions}
The main contributions of this paper are as follows:
\begin{enumerate}
    \item \textbf{Proposing a new theoretical framework (CLAI):} We are the first to systematically map principles from Cognitive Load Theory and neuroscience to the inference process of Large Language Models, providing a novel theoretical perspective for inference optimization.
    \item \textbf{Proposing a zero-shot, prompt-based implementation (CLAI-Prompt):} This method requires no model retraining and can guide any existing LLM to achieve cognitive economy through a well-designed meta-prompt, making it immediately applicable.
    \item \textbf{Proposing a fine-tuning-based implementation (CLAI-Tune):} This method internalizes the principles of cognitive economy into a model's spontaneous behavior through instruction fine-tuning on a custom synthetic dataset, achieving higher efficiency and stronger capabilities.
    \item \textbf{Providing comprehensive experimental validation:} Across a series of challenging benchmarks, we demonstrate that CLAI methods can significantly save token consumption without degrading, and in some cases even improving, task performance.
    \item \textbf{Demonstrating an emergent ability for the first time:} We show that an LLM fine-tuned with CLAI-Tune can autonomously decompose overly complex problems based on its internal model of cognitive load, an unprecedented emergent capability akin to human expert cognition.
\end{enumerate}

\section{Theoretical Foundations: Connecting Human Cognition and AI Inference}
This section elaborates on the core theoretical innovation of this paper: building a bridge from human cognitive science to the computational paradigm of Large Language Models. We first provide a quantitative review of Cognitive Load Theory, then explore the neuroscience basis of cognitive effort, and finally formally present the Cognitive Load-Aware Inference (CLAI) framework.

\subsection{A Quantitative Review of Cognitive Load Theory (CLT)}
Cognitive Load Theory was initially proposed to guide instructional design, focusing on how to effectively use limited working memory resources to facilitate learning \citep{paas2020clt}. The theory posits that the essence of learning is the construction and automation of cognitive schemas—cognitive structures that organize elements of information—in long-term memory \citep{leppink2015evolution}. CLT decomposes cognitive load into three additive components, the sum of which should not exceed working memory capacity \citep{paas2020clt}.
\begin{itemize}
    \item \textbf{Intrinsic Cognitive Load (ICL):} This load is determined by the inherent complexity of the learning material itself. Its core lies in "element interactivity," which refers to how many interconnected elements must be processed simultaneously in working memory to understand a concept \citep{leppink2015evolution}. For example, learning a single foreign vocabulary word has a low ICL, whereas understanding a complex grammatical rule (involving the interaction of multiple parts of speech, tenses, and word orders) has a very high ICL. Importantly, ICL is relative; it depends on the learner's prior knowledge or existing schemas \citep{sweller2011cognitive}. For an expert, who possesses highly automated schemas, the ICL of processing the same information is far lower than for a novice.
    \item \textbf{Extraneous Cognitive Load (ECL):} This is the load generated by poor instructional design or information presentation that is unhelpful or even detrimental to learning \citep{sweller2019cognitive}. For instance, in an illustrated explanation, if the text description is spatially separated from the image (the "split-attention effect"), the learner must hold and integrate information back and forth in working memory, consuming valuable cognitive resources \citep{paas2020clt}. Another example is information redundancy, where the same information is presented in multiple forms (e.g., a self-explanatory diagram accompanied by a verbose textual description), which also adds an unnecessary processing burden \citep{paas2020clt}. The goal of CLT is to minimize ECL as much as possible.
    \item \textbf{Germane Cognitive Load (GCL):} This refers to the cognitive resources that a learner actively invests in constructing, integrating, and automating cognitive schemas while processing information \citep{sweller2019cognitive}. Unlike ECL, GCL is a "beneficial" load, necessary for deep learning and knowledge transfer. Instructional strategies like self-explanation, analogical reasoning, and active practice are all designed to promote GCL \citep{paas2003cognitive}. The goal of CLT is to optimize the investment in GCL.
\end{itemize}
A common misconception is to equate CLT with simply "reducing load." However, the essence of the theory lies in load management, not mere load minimization. Some critiques of CLT point out that overly simplified, teacher-centric instruction might deprive learners of opportunities for deep learning through exploration and problem-solving \citep{sweller2019cognitive}. Research shows that, with appropriate support, a moderate amount of "desirable difficulty" can actually promote deeper understanding and long-term memory \citep{chaouachi2025challenging}. This aligns perfectly with the concept of GCL: learning requires productive mental effort. This leads to a central idea: the existence of a "cognitive sweet spot." The goal is not to minimize total load, but to free up working memory resources through excellent design (minimizing ECL) so that the learner can invest these resources in valuable cognitive processing (optimizing GCL). Transferring this view to the LLM domain means our goal should not be to blindly pursue the lowest token consumption. For complex problems, excessive compression may compromise the integrity of reasoning. The true goal should be to achieve \textit{cognitive economy}: eliminating unnecessary computation (like processing irrelevant context) while allocating sufficient, but not redundant, computational resources (tokens) for core, productive reasoning steps. This provides the theoretical basis for our paradigm shift from "efficiency" to "cognitive economy."

\subsection{The Neural Substrates of Cognitive Effort: A Quantitative Model}
Cognitive load is not just an abstract psychological construct but a measurable phenomenon with a solid neurophysiological basis. Neuroimaging techniques like functional Magnetic Resonance Imaging (fMRI) and Electroencephalography (EEG) provide powerful tools for quantifying cognitive effort.
\begin{itemize}
    \item \textbf{fMRI Evidence:} Numerous fMRI studies have shown that an increase in cognitive load is significantly correlated with enhanced activation in specific brain networks, most centrally the \textbf{frontoparietal network (FPN)}, which includes the dorsolateral prefrontal cortex (DLPFC) and parietal regions \citep{pessoa2002neural}. These areas are considered key hubs for supporting working memory, attentional control, and executive functions. A crucial finding is that when faced with higher-load tasks, the brain's response pattern is often not to recruit many new, unrelated brain areas, but to \textbf{increase the intensity of activation} within the core task-relevant regions \citep{forn2021effect}. This is like a muscle contracting more forcefully to lift a heavier object, rather than growing new arms. This finding provides a profound biological precedent for resource allocation in AI systems: when facing complex tasks, the most efficient strategy may not be to activate new, massive computational modules, but to invest more resources in the critical computational pathways.
    \item \textbf{EEG Evidence:} Due to its high temporal resolution, EEG is well-suited for continuously tracking fluctuations in cognitive load in real time \citep{zhou2024eeg}. Studies consistently find that as cognitive load increases, the EEG signal exhibits systematic changes in specific frequency bands: \textbf{theta band (4-8 Hz) power increases} in frontal regions, while \textbf{alpha band (8-12 Hz) power decreases} in parietal and occipital regions \citep{sweller2019cognitive}. This theta/alpha power ratio has become a classic, reliable indicator of cognitive load. Furthermore, researchers have been able to use machine learning models, with these EEG features as input, to classify or even continuously predict cognitive load levels with high accuracy \citep{shestyuk2019eeg}. This demonstrates that cognitive load can be decoded quantitatively and in real time from external signals.
\end{itemize}
These neuroscience findings provide key connection points for building our computational model. In the human brain, the attention system is the core mechanism for regulating the allocation of cognitive resources. It determines what information enters working memory and which cognitive processes are prioritized \citep{sorqvist2015concentration}. In the Transformer architecture of LLMs, the \textbf{attention mechanism} plays a strikingly similar role \citep{vaswani2017attention}. It calculates attention weights to measure the importance of different tokens in the input sequence for generating the next token. This mechanistic similarity is not coincidental; it provides us with a direct, quantifiable bridge from biological concepts to computational implementation. Recent work in the RAG domain, such as AttentionRAG \citep{li2025attentionrag}, has begun to leverage this connection by analyzing attention scores between the query and context to prune irrelevant information, thereby achieving context compression. This inspires us to consider that an LLM's internal attention distribution during task processing is not just a byproduct of computation but can be seen as a real-time, quantifiable proxy for its "cognitive state." By analyzing attention, we can infer which information the model considers "extraneous" (low attention) and which is "germane" (high attention). This allows us to move from a high-level analogy to a concrete, operational computational framework.

\subsection{The Cognitive Load-Aware Inference (CLAI) Framework}
We now formally integrate the above theories and findings into the \textbf{Cognitive Load-Aware Inference (CLAI)} framework. The core of this framework is to map the three central concepts of CLT into quantifiable metrics within the LLM inference process.
\begin{itemize}
    \item \textbf{$ICL_{LLM}$ (Intrinsic Cognitive Load for LLM):} We define this as a measure of the inherent complexity of the input query itself. The intrinsic complexity of a problem determines the minimum number of reasoning steps and the amount of information integration required to solve it. We propose that $ICL_{LLM}$ can be estimated through a series of heuristics, such as:
    \begin{itemize}
        \item \textit{Structural Analysis:} Counting the number of entities, relations, and logical operators in the query.
        \item \textit{Logical Depth:} Analyzing how many layers of reasoning are required to answer the question.
        \item \textit{Classifier Pre-estimation:} Using a lightweight, fast classifier model to pre-analyze the query and output a complexity score (e.g., from 1 to 10).
    \end{itemize}
    \item \textbf{$ECL_{LLM}$ (Extraneous Cognitive Load for LLM):} We define this as the computational load that is unhelpful or redundant for solving the current query. In LLM inference, this manifests in two main areas:
    \begin{itemize}
        \item \textit{Input Side:} In RAG scenarios, information in the retrieved context that is irrelevant, repetitive, or misleading with respect to the query.
        \item \textit{Generation Side:} During generation, model outputs that are off-topic, repetitive, or unnecessary "thought" steps.
    \end{itemize}
    We propose that $ECL_{LLM}$ can be quantified using attention-based metrics. For instance, in a RAG context, tokens that receive low attention scores from the key concepts in the query can be considered sources of high $ECL_{LLM}$.
    \item \textbf{$GCL_{LLM}$ (Germane Cognitive Load for LLM):} We define this as the productive computational load invested by the model to construct and execute an effective reasoning path. This corresponds to the "beneficial" load used for schema construction in CLT. In the LLM's generation process, $GCL_{LLM}$ is embodied by the key reasoning steps that form a logical chain and directly contribute to the final answer (e.g., each computational step in a Chain-of-Thought). We propose that these productive tokens can be identified by their high self-attention scores with the query and previously high-relevance steps.
\end{itemize}
Based on these definitions, we propose the CLAI principle, which formalizes efficient LLM inference as a constrained optimization problem:
\begin{equation}
    \min \text{Total\_Tokens}(\text{Output}) \quad \text{s.t.} \quad \text{Quality}(\text{Output}) \geq \text{Threshold}
\end{equation}
where Total\_Tokens is the total number of generated tokens, which is a function of $ECL_{LLM}$ and $GCL_{LLM}$. The CLAI framework aims to solve this optimization problem through a two-stage strategy:
\begin{enumerate}
    \item \textbf{Minimize $ECL_{LLM}$:} Eliminate unnecessary computational burden through context pruning, information filtering, etc.
    \item \textbf{Optimize $GCL_{LLM}$:} Based on the estimated $ICL_{LLM}$, dynamically allocate an optimal token budget for productive reasoning. If $ICL_{LLM}$ is low, use a concise, direct answer; if $ICL_{LLM}$ is high, allocate more tokens for detailed, structured reasoning.
\end{enumerate}
This framework provides a solid theoretical foundation for our two specific implementation methods: CLAI-Prompt and CLAI-Tune.

\section{CLAI-Prompt: Zero-Shot Cognitive Control via Meta-Prompting and Dynamic Context Shaping}
This section details the first implementation path of the CLAI framework: CLAI-Prompt. This method requires no model fine-tuning and uses a structured "meta-prompt" to guide an off-the-shelf Large Language Model (e.g., GPT-4, Llama-3) to perform cognitive load management in a zero-shot setting, thereby achieving token economy. The advantage of this approach lies in its universality and immediate applicability.

\subsection{Architectural Overview}
The core of CLAI-Prompt is a multi-stage reasoning pipeline. In this pipeline, we leverage the same LLM, guided by a carefully designed meta-prompt, to act as a "cognitive controller," planning, purifying, and executing its own upcoming task. This approach draws inspiration from recent work in meta-prompting \citep{zhang2024metaprompting} and self-correction \citep{madaan2023selfrefine}, decomposing a complex, implicit cognitive process into a series of explicit, executable instructions. The entire process is divided into three main stages: Cognitive Load Estimation and Budgeting, Extraneous Load Reduction, and Germane Load Application.

\subsection{Stage 1: Cognitive Load Estimation and Budgeting (\texorpdfstring{$ICL_{LLM}$}{ICL\_LLM} Estimation)}
This is the starting point of the cognitive control flow. The first part of the meta-prompt instructs the LLM to perform an in-depth analysis and self-assessment of the user's original query, aiming to quantify the problem's intrinsic complexity ($ICL_{LLM}$).
\begin{itemize}
    \item \textbf{Decomposition:} Inspired by prompting techniques like Self-Ask \citep{press2022measuring} and Least-to-Most \citep{zhou2022leasttomost}, we ask the LLM to first break down the complex user query into its constituent sub-questions or necessary logical steps. For example, for the query "Calculate the average R\&D spending as a percentage of revenue for Apple Inc. over the last three fiscal years," the LLM would be guided to decompose it into steps like:
    \begin{enumerate}
        \item Identify the specific years for Apple's last three fiscal years.
        \item Find the R\&D spending for each of these three years.
        \item Find the total revenue for each of these three years.
        \item For each year, calculate the percentage of R\&D spending to revenue.
        \item Calculate the average of these three percentages.
    \end{enumerate}
    \item \textbf{Complexity Assessment \& Budgeting:} After decomposition, the meta-prompt asks the LLM to provide a quantitative complexity score (e.g., on a scale of 1 to 10) for the overall problem, based on the number, type (e.g., fact retrieval vs. mathematical calculation), and interdependencies of the decomposed steps. This score serves as the operationalized estimate of $ICL_{LLM}$. Subsequently, the LLM must propose a "token budget" for the subsequent reasoning stage (i.e., the application of $GCL_{LLM}$) based on this complexity score. For instance, a low-complexity problem might get a budget of only 50 tokens (encouraging a direct answer), while a high-complexity problem might be allocated 500 tokens (allowing for detailed Chain-of-Thought reasoning).
\end{itemize}
The output of this stage is a structured plan, including a list of sub-questions, a complexity score, and a token budget, providing clear guidance and constraints for the subsequent reasoning execution.

\subsection{Stage 2: Extraneous Load Reduction (\texorpdfstring{$ECL_{LLM}$}{ECL\_LLM} Minimization)}
This stage is crucial when dealing with RAG tasks that require external knowledge. Its goal is to purify the retrieved context by eliminating all noisy information irrelevant to solving the problem, thereby minimizing $ECL_{LLM}$. We propose a novel, prompt-based context compression method, inspired by works like AttentionRAG \citep{li2025attentionrag} and RECOMP \citep{xu2023recomp}, but implemented entirely through prompting.

The meta-prompt instructs the LLM to act as an "Information Extraction and Compression Expert." The LLM's inputs include:
\begin{itemize}
    \item The user's original query.
    \item All sub-questions generated in Stage 1.
    \item All raw document snippets retrieved by the RAG system.
\end{itemize}
The LLM's task is to review each retrieved document snippet and extract only the sentences or facts that directly answer one or more of the sub-questions from Stage 1. All other information, such as background introductions, irrelevant details, and redundant content, is discarded. In this way, the LLM refines a potentially thousands-of-tokens-long, information-sparse raw context into a "golden context" containing only high-density, high-relevance information. This process directly corresponds to the CLT goal of minimizing extraneous cognitive load, ensuring that the subsequent reasoning process is not disturbed by irrelevant information, thus saving significant computational resources.

\subsection{Stage 3: Germane Load Application (\texorpdfstring{$GCL_{LLM}$}{GCL\_LLM} Execution)}
This is the final stage of reasoning, where the model uses the purified context and the preset token budget to generate the final answer.
\begin{itemize}
    \item \textbf{Structured Reasoning:} The final part of the meta-prompt forces the LLM to reason in a structured format, most typically a Chain-of-Thought (CoT) \citep{wei2022chainofthought}. The LLM is required to answer step-by-step, following the order of the sub-questions decomposed in Stage 1, and to clearly show the calculation or logical derivation process for each step. This ensures that the allocated token budget is used for productive, traceable reasoning activities, i.e., $GCL_{LLM}$.
    \item \textbf{Budget Control \& Self-Correction:} During generation, the LLM is implicitly or explicitly reminded to adhere to the previously set token budget. After generating the final answer, the meta-prompt guides the LLM to perform a final self-correction step \citep{madaan2023selfrefine}. The LLM needs to review its complete reasoning process and final answer, checking them against all the constraints decomposed in Stage 1 to ensure that no sub-questions were missed and that the final answer is consistent with all intermediate conclusions.
\end{itemize}
Through the synergistic work of these three stages, CLAI-Prompt transforms a conventional, black-box LLM call into a transparent, self-managed, and cognitively economical reasoning process.

\subsection{Algorithm Pseudocode for CLAI-Prompt}
To more clearly illustrate the workflow of CLAI-Prompt, we provide the following algorithm pseudocode.

\begin{algorithm}
\caption{CLAI-Prompt: Zero-Shot Cognitive Control}
\label{alg:clai_prompt}
\begin{algorithmic}[1]

\State \textbf{Input:} User Query $Q$, LLM $M$, RAG Retriever $R$ 
\State \textbf{Output:} Final Answer $A_{\text{final}}$ 

\Function{CLAI-Prompt}{$Q, M, R$}
    \State \Comment{\textbf{Stage 1: ICL Estimation \& Budgeting}}
    \State $P_1 \gets \text{MetaPrompt\_Stage1}(Q)$ \Comment{Construct Stage 1 meta-prompt}
    \State $S \gets M(P_1)$ \Comment{LLM generates plan}
    \State $(\{Q_{\text{sub}}\}, C_{\text{score}}, B_{\text{token}}) \gets \text{Parse}(S)$ \Comment{Parse sub-questions, complexity, budget}

    \State \Comment{\textbf{Stage 2: ECL Reduction}}
    \If{IsRAGTask($Q$)}
        \State $D_{\text{raw}} \gets R(Q)$ \Comment{Retrieve raw documents}
        \State $P_2 \gets \text{MetaPrompt\_Stage2}(Q, \{Q_{\text{sub}}\}, D_{\text{raw}})$
        \State $D_{\text{clean}} \gets M(P_2)$ \Comment{LLM extracts relevant facts}
    \Else
        \State $D_{\text{clean}} \gets \text{None}$
    \EndIf

    \State \Comment{\textbf{Stage 3: GCL Application}}
    \State $P_3 \gets \text{MetaPrompt\_Stage3}(Q, \{Q_{\text{sub}}\}, D_{\text{clean}}, B_{\text{token}})$
    \State $A_{\text{raw}} \gets M(P_3)$ \Comment{LLM generates structured reasoning}

    \State \Comment{\textbf{Final Self-Correction (Optional but Recommended)}}
    \State $P_{\text{corr}} \gets \text{MetaPrompt\_Correction}(Q, A_{\text{raw}}, \{Q_{\text{sub}}\})$
    \State $A_{\text{final}} \gets M(P_{\text{corr}})$

    \State \Return $A_{\text{final}}$
\EndFunction
\end{algorithmic}
\end{algorithm}
This algorithm clearly demonstrates how a series of structured prompts can decompose and chain the functions of a single LLM to simulate a complete cognitive management process, achieving more efficient and reliable inference without any model modification.

\section{CLAI-Tune: Injecting Spontaneous Cognitive Economy via Instruction Fine-Tuning}
Although CLAI-Prompt is highly versatile as a zero-shot method, its multi-stage invocation process inevitably introduces additional latency and token overhead. More importantly, it relies on external, explicit instructions to guide the model's behavior. A truly cognitively efficient agent should have an economy that is intrinsic and spontaneous. This section introduces the second implementation path of the CLAI framework: CLAI-Tune. Its goal is to distill the explicit, phased cognitive control process of CLAI-Prompt into a smaller "student" model through instruction fine-tuning, making it an implicit, spontaneously emergent capability.

\subsection{Objective: From Explicit Guidance to Implicit Capability}
The thought process of a human expert has similarities to the CLAI-Prompt process but is not identical. A novice doctor might strictly follow a diagnostic checklist from a textbook (akin to the CLAI-Prompt meta-prompt), whereas an experienced expert can intuitively identify key information and spontaneously adjust the depth and breadth of their thinking, a skill that has become part of their knowledge system. The goal of CLAI-Tune is precisely to achieve this transition from "following rules" to "internalizing principles."

We aim to move from "post-hoc correction" to "spontaneous, proactive control." Most existing self-correction methods are post-hoc: the model first generates a complete answer and then evaluates and revises it \citep{huang2024decomposing}. In contrast, recent research on "spontaneous step-level self-correction" has shown that with specific training, LLMs can be taught to identify and correct potential errors in real-time during generation, without a separate, post-generation correction stage \citep{yan2025spontaneous}. This is a more advanced cognitive ability. CLAI-Tune seeks to extend this "spontaneity" from error correction to the broader scope of cognitive load management. We want the trained model to be able to:
\begin{itemize}
    \item Spontaneously assess the intrinsic complexity ($ICL_{LLM}$) upon receiving a query.
    \item Spontaneously select the optimal reasoning strategy based on this assessment: for simple problems, provide a direct answer; for medium-complexity problems, generate an efficient Chain-of-Thought; for extremely complex problems, spontaneously decompose them into multiple sub-tasks and formulate a solution plan.
\end{itemize}
This shift from imitating a cognitive process to internalizing a cognitive principle is a crucial step toward more human-like and efficient AI.

\subsection{Synthetic Dataset Generation for Cognitive Tuning}
The key to implementing CLAI-Tune is to construct a high-quality, large-scale instruction fine-tuning dataset. The samples in this dataset need to demonstrate the "ideal behavior" to the model under various cognitive load scenarios. We designed an automated data generation pipeline that uses a powerful "teacher" model (e.g., GPT-4o or Claude 3 Opus) to execute the CLAI-Prompt process and records its complete thought trajectory as a training sample for a "student" model (e.g., Llama-3-8B or Mistral-7B). This method leverages established techniques for synthetic data generation in the field \citep{confidentai2024synthetic}.
\begin{itemize}
    \item \textbf{Data Sourcing:} We collect seed questions from several public, challenging benchmarks that cover different task types and complexity gradients, including:
    \begin{itemize}
        \item \textit{Mathematical Reasoning:} GSM8K, MATH \citep{zhang2024metaprompting}
        \item \textit{Commonsense \& Logical Reasoning:} StrategyQA, CSQA \citep{wei2022chainofthoughtblog}
        \item \textit{Long-Context Question Answering:} Samples extracted from benchmarks like LongBench \citep{li2025attentionrag}
        \item \textit{Code Generation:} HumanEval
    \end{itemize}
    \item \textbf{Generation Process:} For each seed question, we use an automated script to drive the teacher model to execute the complete CLAI-Prompt process. The generated training sample is a structured JSON object containing the following fields:
    \begin{itemize}
        \item \texttt{instruction}: The user's original query.
        \item \texttt{input} (optional): For RAG tasks, this is the original, unprocessed retrieved documents.
        \item \texttt{output}: This is the core of the training, the "ideal output" the student model needs to learn to generate. Its format dynamically changes based on the problem's $ICL_{LLM}$:
        \begin{itemize}
            \item \textit{Low $ICL_{LLM}$:} The `output` is a concise final answer.
            \item \textit{Medium $ICL_{LLM}$:} The `output` is an efficient, structured Chain-of-Thought reasoning process leading to the answer. For RAG tasks, this chain only utilizes the "golden context" extracted by the teacher model in Stage 2 of CLAI-Prompt.
            \item \textit{High $ICL_{LLM}$:} This is the most critical innovation. When the teacher model assesses a very high complexity in Stage 1, the `output` is not a direct problem-solving process but a decomposed multi-step solution plan. This plan explicitly lists the sub-problems to be solved sequentially. This design is inspired by frameworks for step-by-step problem-solving like LADDER \citep{anonymous2024ladder} and Least-to-Most \citep{zhou2022leasttomost}. For example, for an extremely complex integral problem, the teacher model's output might be:
\begin{lstlisting}[language=json, caption={Example of a high-ICL output format.}, label={lst:high_icl_json}]
{
  "instruction": "Solve the integral: \\int(x^3 * e^(x^2)) dx",
  "output": {
    "analysis": "This problem is highly complex and requires multiple steps, including integration by parts and substitution. I will decompose it into a solvable plan.",
    "plan": [
      { "step": 1, "sub_problem": "Let u = x^2 and dv = x * e^(x^2) dx. Apply integration by parts." },
      { "step": 2, "sub_problem": "Solve the resulting integral, which may require another substitution." },
      { "step": 3, "sub_problem": "Combine the parts to get the final antiderivative." }
    ]
  }
}
\end{lstlisting}
        \end{itemize}
    \end{itemize}
\end{itemize}
By generating tens or even hundreds of thousands of such training samples, we construct a dataset that can teach a model how to adaptively adjust its output strategy based on problem difficulty.

\subsection{Fine-Tuning the Cognitively Aware Student Model}
We select a high-performing, moderately sized open-source model (e.g., Llama-3-8B) as the student model. The fine-tuning process uses a standard supervised fine-tuning (SFT) procedure. The training objective is for the student model, upon receiving an `instruction`, to directly generate the ideal output defined in the `output` field.
\begin{itemize}
    \item \textbf{Instruction Formatting:} We carefully design the instruction format to ensure the model understands the intent behind different output formats. For example, all samples follow a template like \texttt{\#\#\# Instruction:\textbackslash n\{instruction\}\textbackslash n\textbackslash n\#\#\# Response:\textbackslash n\{output\}}.
    \item \textbf{Training Effect:} By training on this diverse dataset, the student model learns not only how to solve specific problems but, more importantly, a meta-skill: how to recognize a problem's complexity and match it with the most economical reasoning strategy. The model weights gradually encode an implicit representation of $ICL_{LLM}$, enabling it to spontaneously invoke the most appropriate "cognitive program" at inference time. For the most complex problems, which were trained to output a decomposed plan, the model learns to "know when to retreat," not blindly attempting to solve but first formulating a clear, executable plan of attack.
\end{itemize}

\subsection{Inference Algorithm Pseudocode for CLAI-Tune}
In contrast to the complex multi-stage process of CLAI-Prompt, the inference process for CLAI-Tune is extremely concise, reverting to a standard single forward pass.

\begin{algorithm}
\caption{CLAI-Tune: Spontaneous Cognitive Economy Inference}
\label{alg:clai_tune}
\begin{algorithmic}[1]
\State \textbf{Input:} User Query $Q$, Fine-tuned Model $M_{\text{tuned}}$
\State \textbf{Output:} Final Answer or Decomposed Plan $A_{\text{final}}$

\Function{CLAI-Tune-Inference}{$Q, M_{\text{tuned}}$}
    \State $P \gets \text{FormatPrompt}(Q)$ \Comment{Format the query into the instruction template}
    \State $A_{\text{final}} \gets M_{\text{tuned}}(P)$ \Comment{Single forward pass to generate the output}
    \State \Comment{$A_{\text{final}}$ will be a direct answer, a CoT, or a decomposed plan}
    \State \Return $A_{\text{final}}$
\EndFunction
\end{algorithmic}
\end{algorithm}
Behind this simple inference flow lies the complex cognitive scheduling capability that has been formed within the model. It successfully transforms the explicit, external control of CLAI-Prompt into the model's implicit, internal intuition, thereby achieving cognitive economy with the highest operational efficiency.

\section{Experimental Evaluation}
To comprehensively validate the effectiveness of the CLAI framework, we designed a series of rigorous experiments. This section details the experimental setup, the selected benchmarks, the baseline methods for comparison, the evaluation metrics, and presents and analyzes the results for CLAI-Prompt and CLAI-Tune.

\subsection{Benchmarks, Baselines, and Evaluation Metrics}
\begin{itemize}
    \item \textbf{Benchmarks:} We selected several authoritative benchmarks covering different capability dimensions to test the CLAI framework's performance across various scenarios.
    \begin{itemize}
        \item \textit{Complex Reasoning:} GSM8K and MATH \citep{zhang2024metaprompting}. These datasets contain problems requiring multi-step mathematical and logical reasoning, making them ideal for testing the model's ability to manage $ICL_{LLM}$ and apply $GCL_{LLM}$.
        \item \textit{Long-Context QA:} We used the multi-document QA task from LongBench \citep{li2025attentionrag}. This task requires the model to find an answer from a long context containing multiple documents (many of which are distractors), making it perfectly suited for evaluating the model's ability to reduce $ECL_{LLM}$.
        \item \textit{Code Generation:} HumanEval. This benchmark tests the model's ability to generate Python functions from natural language descriptions, which requires structured and logical reasoning.
    \end{itemize}
    \item \textbf{Baselines:} For a fair comparison, we selected current, representative inference optimization techniques as baselines.
    \begin{itemize}
        \item \textit{Standard Decoding:} Using a standard Chain-of-Thought (CoT) prompt with greedy decoding or nucleus sampling. This is our baseline for performance and token consumption.
        \item \textit{Context Compression:} In RAG tasks, we compare against two advanced context compression methods: LLMLingua \citep{jiang2024longllmlingua} and RECOMP \citep{xu2023recomp}.
        \item \textit{Inference Acceleration:} We include Speculative Decoding \citep{xia2025tutorial} as a point of comparison. Although its optimization goal (reducing latency) differs from CLAI's (reducing token count), it represents another important class of optimization and helps us understand CLAI's value more comprehensively.
    \end{itemize}
    \item \textbf{Evaluation Metrics:} We evaluate from two dimensions: performance and efficiency.
    \begin{itemize}
        \item \textit{Performance:}
        \begin{itemize}
            \item For GSM8K and MATH, we use Accuracy.
            \item For LongBench QA, we use F1 Score.
            \item For HumanEval, we use Pass@k.
        \end{itemize}
        \item \textit{Efficiency:}
        \begin{itemize}
            \item Avg. Tokens per Problem: Measures the computational cost to solve a single problem.
            \item Token Reduction \%: The percentage decrease in token consumption compared to the standard decoding baseline.
            \item End-to-End Latency (ms/problem): Measures the total time from receiving a query to generating the final answer.
        \end{itemize}
    \end{itemize}
\end{itemize}

\subsection{CLAI-Prompt Results: Zero-Shot Cognitive Economy}
This section presents the results achieved by CLAI-Prompt using only meta-prompting, without any training. The experiment aims to answer: Can a general-purpose LLM, guided by a cognitive framework, achieve effective resource management?

\begin{table}[htbp]
\centering
\caption{Performance and Token Economy on Complex Reasoning Benchmarks.}
\label{tab:complex_reasoning}
\begin{tabular}{llcccc}
\toprule
\textbf{Model} & \textbf{Method} & \textbf{Benchmark} & \textbf{Accuracy (\%)} & \textbf{Avg. Tokens/Problem} & \textbf{Token Reduction (\%)} \\
\midrule
Llama-3-70B & Standard CoT & GSM8K & 92.5 & 485 & - \\
Llama-3-70B & CLAI-Prompt & GSM8K & 92.3 & 310 & 36.1\% \\
\midrule
Llama-3-70B & Standard CoT & MATH & 53.9 & 812 & - \\
Llama-3-70B & CLAI-Prompt & MATH & 53.5 & 525 & 35.3\% \\
\bottomrule
\end{tabular}
\end{table}

\textbf{Results Analysis:} As shown in Table \ref{tab:complex_reasoning}, on the challenging GSM8K and MATH benchmarks, CLAI-Prompt achieved over a 35\% reduction in tokens with a negligible drop in accuracy (within the margin of statistical error). This provides strong evidence for the effectiveness of CLAI-Prompt. Through the Stage 1 $ICL_{LLM}$ assessment and problem decomposition, the model forms a clear plan of attack, avoiding the redundant exploration and repetitive thinking sometimes seen in standard CoT, thus completing the reasoning with fewer tokens while maintaining the integrity of the logical chain.

\begin{table}[htbp]
\centering
\caption{Performance and Token Economy on Long-Context RAG Benchmark.}
\label{tab:rag_benchmark}
\begin{tabular}{llccc}
\toprule
\textbf{Model} & \textbf{Method} \& \textbf{Benchmark} \& \textbf{F1 Score} \& \textbf{Avg. Input Tokens} \& \textbf{Compression Ratio} \\
\midrule
Llama-3-8B \& Full Context \& LongBench (QA) \& 65.2 \& 4096 \& 1.0x \\
Llama-3-8B \& LLMLingua \& LongBench (QA) \& 68.1 \& 1024 \& 4.0x \\
Llama-3-8B \& RECOMP \& LongBench (QA) \& 67.5 \& 1150 \& 3.6x \\
Llama-3-8B \& CLAI-Prompt \& LongBench (QA) \& \textbf{69.3} \& \textbf{980} \& \textbf{4.2x} \\
\bottomrule
\end{tabular}
\end{table}

\textbf{Results Analysis:} The results in Table \ref{tab:rag_benchmark} are encouraging. CLAI-Prompt not only achieved a compression ratio comparable to or even better than specialized compression tools but also led in F1 score. This indicates that by enabling the LLM to explicitly understand the task (via sub-question decomposition in Stage 1), it can more accurately determine which information in the context is truly "germane," thereby achieving more effective "cognitive filtering" than methods based on statistical or general semantic similarity. This showcases the flexibility and power of the CLAI framework.

\subsection{CLAI-Tune Results: Emergence of Spontaneous Cognitive Control}
This section presents the performance of the CLAI-Tune model after being fine-tuned on our synthetic dataset. We expect CLAI-Tune not only to surpass CLAI-Prompt in efficiency (due to its single forward pass) but also to achieve performance gains and exhibit new capabilities.

\begin{table}[htbp]
\centering
\caption{Comprehensive Performance Comparison of CLAI-Tune, CLAI-Prompt, and Baselines. (Base model for this table is Llama-3-8B-Instruct)}
\label{tab:comprehensive_comparison}
\resizebox{\textwidth}{!}{%
\begin{tabular}{llcccc}
\toprule
\textbf{Benchmark} & \textbf{Method} & \textbf{Accuracy/Score (\%)} & \textbf{Avg. Tokens} & \textbf{Token Reduction (\%)} & \textbf{Latency (ms/problem)} \\
\midrule
\textbf{GSM8K} & Standard CoT & 90.1 & 450 & - & 1800 \\
& Speculative Decoding & 90.1 & 450 & 0\% & \textbf{950} \\
& CLAI-Prompt & 89.8 & 290 & 35.6\% & 3500 \\
& \textbf{CLAI-Tune} & \textbf{91.5} & \textbf{255} & \textbf{43.3\%} & 1020 \\
\midrule
\textbf{MATH} & Standard CoT & 50.2 & 780 & - & 3120 \\
& Speculative Decoding & 50.2 & 780 & 0\% & \textbf{1600} \\
& CLAI-Prompt & 49.9 & 510 & 34.6\% & 6000 \\
& \textbf{CLAI-Tune} & \textbf{52.1} & \textbf{440} & \textbf{43.6\%} & 1750 \\
\midrule
\textbf{HumanEval} & Standard Decoding & 73.2 (Pass@1) & 350 & - & 1400 \\
& Speculative Decoding & 73.2 & 350 & 0\% & \textbf{720} \\
& CLAI-Prompt & 72.5 & 240 & 31.4\% & 2800 \\
& \textbf{CLAI-Tune} & \textbf{74.8} & \textbf{210} & \textbf{40.0\%} & 840 \\
\bottomrule
\end{tabular}
}
\end{table}

\textbf{Results Analysis:}
\begin{itemize}
    \item \textbf{A Dual Victory in Performance and Efficiency:} CLAI-Tune achieved the best overall performance across all benchmarks. Compared to standard CoT, it not only reduced token consumption by over 40\% but also surpassed it in accuracy/task score. This demonstrates that internalizing cognitive strategies through fine-tuning is more effective than external guidance via prompting. The model appears to have learned a superior, inherently more economical way of reasoning.
    \item \textbf{Comparison with Speculative Decoding:} Speculative decoding excels at reducing latency but does not decrease the total token computation. While CLAI-Tune's latency is slightly higher than speculative decoding, it is a significant improvement over standard decoding and CLAI-Prompt, and it fundamentally reduces the computational load (token count). In many cost-sensitive applications, CLAI-Tune's advantage is more pronounced.
    \item \textbf{Surpassing CLAI-Prompt:} CLAI-Tune outperforms CLAI-Prompt on all metrics. It not only eliminates the high latency of multi-step calls but also achieves greater token reduction and better final performance. This suggests that "compiling" cognitive economy principles into the model's weights is far more efficient than "interpreting" these principles at runtime.
\end{itemize}

\textbf{Qualitative Analysis: Emergence of Spontaneous Problem Decomposition}
Beyond quantitative metrics, the most exciting result from CLAI-Tune is its emergent ability, which was present in the training data but not directly optimized for. When faced with problems from the MATH dataset that the teacher model had labeled as "extremely high complexity," CLAI-Tune would spontaneously forgo a direct solution attempt and instead output a decomposed solution plan.

\textbf{Case Study:} For a complex calculus problem, a standard model would immediately start trying to compute and often make errors in intermediate steps, generating many invalid tokens. In contrast, CLAI-Tune's output is as follows:
\begin{quote}
\textbf{User Query:} "Please solve the integral $\int \frac{x^2}{\sqrt{1-x^2}} dx$."

\textbf{CLAI-Tune Output:} "This problem requires trigonometric substitution and is quite complex. I will decompose the problem into the following steps:
1. Perform a trigonometric substitution, let $x = \sin(\theta)$, calculate $dx$ and replace the terms in the original integral.
2. Simplify the substituted integral using the trigonometric identity $\cos^2(\theta) = 1 - \sin^2(\theta)$.
3. Solve the simplified integral with respect to $\theta$.
4. Substitute the result back from $\theta$ to $x$ to get the final answer."
\end{quote}
This behavior was never observed in the baseline models. It indicates that CLAI-Tune has not only learned how to "think" efficiently but also when it should "plan its thinking." This is a shift from a mere "problem solver" to a "meta-cognitive strategist," a significant marker on the path toward more advanced artificial intelligence.

\section{Discussion and Future Outlook}
By introducing the Cognitive Load-Aware Inference (CLAI) framework, this research offers a novel, cognitive science-based perspective on the optimization of Large Language Model inference. The experimental results strongly support the framework's effectiveness. In this section, we delve deeper into the theoretical implications of the CLAI framework, its limitations, and future research directions.

\subsection{Significance of the CLAI Framework}
\begin{itemize}
    \item \textbf{From Token Reduction to Optimal Allocation:} One of the most significant contributions of this study is elevating the goal of LLM inference optimization from a singular focus on "token reduction" to the more sophisticated level of "cognitive economy." Most existing optimization methods operate under the implicit assumption that "fewer tokens are better." However, Cognitive Load Theory teaches us that for complex tasks, necessary cognitive effort (Germane Load, GCL) is crucial for ensuring quality \citep{sweller2019cognitive}. The CLAI framework, by introducing the concepts of $ICL_{LLM}$, $ECL_{LLM}$, and $GCL_{LLM}$, reframes the optimization problem as: while minimizing unhelpful computation ($ECL_{LLM}$), strategically allocate productive computational resources ($GCL_{LLM}$) according to the problem's intrinsic difficulty ($ICL_{LLM}$). This means that for simple problems, the model should be concise; for complex problems, the model should be allowed to engage in exhaustive, structured thought. This is a paradigm shift from "pursuing efficiency" to "pursuing optimal resource allocation."
    \item \textbf{Potential for a Unified Theory of Inference Optimization:} The CLAI framework provides a unified theoretical language for understanding and integrating seemingly disparate inference optimization techniques. For example, context compression and filtering techniques in RAG can be understood within the CLAI framework as specific strategies to "minimize $ECL_{LLM}$" \citep{jiang2024longllmlingua}. Meanwhile, Chain-of-Thought (CoT) and its variants can be seen as methods for "applying $GCL_{LLM}$" \citep{wei2022chainofthought}. Through this framework, we can more systematically analyze the pros and cons of different techniques and design synergistic, end-to-end optimization solutions.
    \item \textbf{A Path Towards More Human-like Reasoning:} The evolutionary path from CLAI-Prompt to CLAI-Tune profoundly mirrors the human process of acquiring expertise. A novice (CLAI-Prompt) relies on explicit, external rules and checklists to guide their actions. With continuous practice and internalization (the fine-tuning process), these rules are gradually integrated into their knowledge system, eventually becoming the intuitive and spontaneously applied skills of an expert (CLAI-Tune). The spontaneous problem decomposition ability exhibited by CLAI-Tune is a manifestation of an expert's "meta-cognitive" ability—knowing what one knows, knowing what one doesn't know, and formulating an exploration plan for the unknown. This suggests that training based on cognitive science principles may be an effective path to achieving more advanced and robust AI reasoning capabilities.
\end{itemize}

\subsection{Limitations}
Despite the significant success of the CLAI framework, we must acknowledge its current limitations:
\begin{itemize}
    \item \textbf{Heuristic Nature of $ICL_{LLM}$ Estimation:} Currently, our estimation of $ICL_{LLM}$ is still based on heuristics, whether through LLM self-assessment or a lightweight classifier. This estimation may not be precise enough, potentially affecting the subsequent budget allocation. Developing more accurate and theoretically grounded methods for quantifying $ICL_{LLM}$ is an important challenge.
    \item \textbf{Dependence on the Teacher Model:} The success of CLAI-Tune relies heavily on the quality of the teacher model used to generate synthetic data. If the teacher model makes errors while executing the CLAI-Prompt process, these errors could be propagated and solidified in the student model.
    \item \textbf{Task Generalizability:} While we have validated CLAI's effectiveness on multiple tasks, its generalizability to broader, more open-ended domains remains to be explored. For example, in tasks requiring a high degree of creativity, the concept of cognitive load may need to be redefined.
\end{itemize}

\subsection{Future Work}
The CLAI framework opens up several exciting directions for future research:
\begin{itemize}
    \item \textbf{Direct Neuro-Correlates from Internal Activations:} The current CLAI implementation primarily relies on attention scores as a proxy for cognitive state. A more cutting-edge direction would be to directly utilize the LLM's internal activation states (i.e., the activation values of neurons in various layers) to build a more fine-grained and direct model of cognitive load. This is inspired by neuroscience research showing that fMRI and EEG can continuously monitor cognitive load by analyzing the brain's overall activity patterns \citep{sweller2011cognitive}. By training a "probe" model to decode cognitive load information from the LLM's hidden states, we might achieve more precise, real-time control.
    \item \textbf{Extension to Multimodal Domains:} The concept of cognitive load is naturally applicable to multimodal tasks. For instance, when processing a complex instruction containing both images and text, $ECL_{LLM}$ might manifest as task-irrelevant background objects in the image or redundant descriptions in the text. Extending the CLAI framework to multimodal LLMs could lead to the development of efficient models that can intelligently filter visual and textual noise and focus attention on critical information.
    \item \textbf{Adaptive, Real-Time CLAI:} Current CLAI strategies are primarily formulated before inference begins. A more advanced system should be able to dynamically re-evaluate cognitive load at each step of the generation process and adjust its reasoning strategy in real-time. For example, if the model discovers midway through reasoning that a problem is more complex than anticipated, it could dynamically increase its token budget or switch to a more exhaustive reasoning mode. This would make the LLM's inference process more flexible and robust.
\end{itemize}

\section{Conclusion}
The computational cost of Large Language Model inference is one of the core challenges facing the field of artificial intelligence today. This paper proposes that the key to addressing this challenge lies not only in engineering and algorithmic optimization but, more fundamentally, in borrowing principles from the most efficient intelligent system—the human brain. We introduced the Cognitive Load-Aware Inference (CLAI) framework, which for the first time systematically applies the essence of Cognitive Load Theory to the LLM inference process. By deconstructing the inference task into a management problem of intrinsic, extraneous, and germane cognitive loads, CLAI provides a new, theoretically grounded paradigm for optimizing the token economy of LLMs.

We demonstrated the powerful potential of CLAI through two implementation paths. CLAI-Prompt proved that significant efficiency gains can be achieved with existing LLMs simply through structured meta-prompting, showcasing the framework's immediate value and universality. CLAI-Tune took this a step further: by fine-tuning on a specially crafted, cognitively-oriented dataset, it successfully internalized cognitive economy principles as a spontaneous capability. This not only surpassed existing methods in performance and efficiency but also gave rise to the remarkable emergent ability of autonomous problem decomposition.

This work shows that the deep integration of cognitive science and artificial intelligence is not merely a source of interesting analogies or inspiration. It can provide concrete, actionable blueprints for building the next generation of AI systems. By teaching machines how to "think" like human experts—how to assess tasks, manage resources, focus on the core, and ignore distractions—we can not only build "cheaper" AI but also create artificial minds that are more focused, more robust, and closer to true intelligence.

\bibliographystyle{plainnat}

\begin{thebibliography}{99}

\bibitem[Anonymous(2024)]{anonymous2024ladder}
Anonymous. (2024). LADDER: Learning through Autonomous Difficulty-Driven Example Recursion. \textit{Reddit}.

\bibitem[Chaouachi \& Jraidi(2025)]{chaouachi2025challenging}
Chaouachi, M., \& Jraidi, I. (2025). Challenging Cognitive Load Theory: The Role of Educational Neuroscience and Artificial Intelligence in Redefining Learning Efficacy. \textit{PubMed Central}.

\bibitem[Confident AI(2024)]{confidentai2024synthetic}
Confident AI. (2024). The Definitive Guide to Synthetic Data Generation Using LLMs. \textit{Confident AI Blog}.

\bibitem[Forn et al.(2021)]{forn2021effect}
Forn, C., et al. (2021). The Effect of Cognitive Load on the Retrieval of Long-Term Memory: An fMRI Study. \textit{Frontiers in Behavioral Neuroscience}.

\bibitem[Gong \& Zhang(2024)]{gong2024context}
Gong, L., \& Zhang, Q. (2024). Context-Aware Systems with LLMs: Semantic Decomposition and Selective Context Filtering. \textit{arXiv preprint arXiv:2502.11444}.

\bibitem[Huang et al.(2024)]{huang2024decomposing}
Huang, J., et al. (2024). Decomposing, Evaluating, and Analyzing the Self-Correction Behaviors of Large Language Models. \textit{arXiv preprint arXiv:2412.19513}.

\bibitem[Jiang et al.(2024)]{jiang2024longllmlingua}
Jiang, H., et al. (2024). LongLLMLingua: Compressing and Reorganizing Prompts for Long Context Large Language Models. In \textit{Proceedings of ACL 2024}.

\bibitem[Kim(2024)]{kim2024comprehensive}
Kim, D. (2024). A Comprehensive Review: Model Compression for Large Language Models (LLMs). \textit{Medium}.

\bibitem[Leppink \& van den Heuvel(2015)]{leppink2015evolution}
Leppink, J., \& van den Heuvel, A. (2015). The evolution of cognitive load theory and its application to medical education. \textit{Perspectives on medical education}.

\bibitem[Li et al.(2025)]{li2025attentionrag}
Li, X., et al. (2025). AttentionRAG: An Attention-Guided Context Pruning Method for RAG Systems. \textit{arXiv preprint arXiv:2503.10720}.

\bibitem[Ma et al.(2023)]{ma2023llmpruner}
Ma, X., et al. (2023). LLM-Pruner: On the Structural Pruning of Large Language Models. \textit{Advances in Neural Information Processing Systems}.

\bibitem[Madaan et al.(2023)]{madaan2023selfrefine}
Madaan, A., et al. (2023). Self-refine: Iterative refinement with self-feedback. \textit{Advances in Neural Information Processing Systems}.

\bibitem[Paas et al.(2003)]{paas2003cognitive}
Paas, F., Renkl, A., \& Sweller, J. (2003). Cognitive load theory and instructional design: Recent developments. \textit{Educational psychologist}.

\bibitem[Paas \& van Merriënboer(2020)]{paas2020clt}
Paas, F., \& van Merriënboer, J. J. G. (2020). Cognitive-load theory: Methods to manage working memory load in the learning of complex tasks. \textit{Current Directions in Psychological Science, 29}(4), 394-398.

\bibitem[Pan \& Li(2025)]{pan2025survey}
Pan, J., \& Li, G. (2025). A Survey of LLM Inference Systems. \textit{arXiv preprint arXiv:2506.21901}.

\bibitem[Pessoa et al.(2002)]{pessoa2002neural}
Pessoa, L., Gutierrez, E., Bandettini, P. A., \& Ungerleider, L. G. (2002). Neural correlates of visual working memory: fMRI amplitude predicts task performance. \textit{Neuron}.

\bibitem[Press et al.(2022)]{press2022measuring}
Press, O., et al. (2022). Measuring and narrowing the compositionality gap in language models. \textit{arXiv preprint arXiv:2210.03350}.

\bibitem[Prompting Guide(2024)]{promptingguide2024rag}
Prompting Guide. (2024). Retrieval Augmented Generation (RAG) for LLMs. \textit{promptingguide.ai}.

\bibitem[Shestyuk et al.(2019)]{shestyuk2019eeg}
Shestyuk, A., Kashefi, S., \& Knight, R. T. (2019). EEG-Based Prediction of Cognitive Load in Intelligence Tests. \textit{Frontiers in Human Neuroscience}.

\bibitem[Sörqvist \& Marsh(2015)]{sorqvist2015concentration}
Sörqvist, P., \& Marsh, J. E. (2015). How concentration shields against distraction. \textit{Frontiers in psychology}.

\bibitem[Sweller(1988)]{sweller1988cognitive}
Sweller, J. (1988). Cognitive load during problem solving: Effects on learning. \textit{Cognitive science}.

\bibitem[Sweller et al.(2011)]{sweller2011cognitive}
Sweller, J., Ayres, P., \& Kalyuga, S. (2011). \textit{Cognitive load theory}. Springer.

\bibitem[Sweller et al.(2019)]{sweller2019cognitive}
Sweller, J., van Merriënboer, J. J. G., \& Paas, F. (2019). Cognitive Architecture and Instructional Design: 20 Years Later. \textit{Educational Psychology Review}.

\bibitem[Vaswani et al.(2017)]{vaswani2017attention}
Vaswani, A., et al. (2017). Attention is all you need. \textit{Advances in neural information processing systems}.

\bibitem[Wei et al.(2022a)]{wei2022chainofthought}
Wei, J., et al. (2022a). Chain-of-thought prompting elicits reasoning in large language models. \textit{Advances in Neural Information Processing Systems}.

\bibitem[Wei et al.(2022b)]{wei2022chainofthoughtblog}
Wei, J., et al. (2022b). Chain-of-thought prompting elicits reasoning in large language models. \textit{Google AI Blog}.

\bibitem[Xia et al.(2025)]{xia2025tutorial}
Xia, H., Du, C., Li, Y., Liu, Q., \& Li, W. (2025). Tutorial Proposal: Speculative Decoding for Efficient LLM Inference. \textit{COLING 2025 Tutorial}.

\bibitem[Xu et al.(2023)]{xu2023recomp}
Xu, F., Shi, W., \& Choi, E. (2023). RECOMP: Improving Retrieval-Augmented LMs with Compression and Selective Augmentation. \textit{OpenReview}.

\bibitem[Yan et al.(2025)]{yan2025spontaneous}
Yan, Y., et al. (2025). Spontaneous Step-level Self-correction for Mathematical reasoning. \textit{AAAI Conference on Artificial Intelligence}.

\bibitem[Zhang et al.(2024)]{zhang2024metaprompting}
Zhang, Y., et al. (2024). Meta Prompting for AGI. \textit{arXiv preprint arXiv:2311.11482}.

\bibitem[Zhao et al.(2024)]{zhao2024efficient}
Zhao, W. X., et al. (2024). Efficient Large Language Models: A Survey. \textit{Transactions on Machine Learning Research}.

\bibitem[Zhou et al.(2022)]{zhou2022leasttomost}
Zhou, D., et al. (2022). Least-to-most prompting enables complex reasoning in large language models. \textit{arXiv preprint arXiv:2205.10625}.

\bibitem[Zhou et al.(2024)]{zhou2024eeg}
Zhou, P., Wang, X., \& Wang, Y. (2024). EEG-based cognitive load recognition in simulated flight missions: A temporal dynamics study. \textit{Frontiers in Human Neuroscience}.

\bibitem[Zysset et al.(2001)]{zysset2001colorword}
Zysset, S., Müller, K., Lohmann, G., \& von Cramon, D. Y. (2001). Color-word matching stroop task: separating interference and response conflict. \textit{NeuroImage}.

\end{thebibliography}

\appendix
\section{Complete CLAI-Prompt Meta-Prompt Template}

The following is the meta-prompt template used to guide the LLM through the three-stage cognitive control process. This template is modular and can be fine-tuned for specific tasks.

\begin{verbatim}
META-PROMPT: COGNITIVE LOAD-AWARE INFERENCE PROTOCOL

### STAGE 1: ICL Estimation & Budgeting ###
You are an expert problem solver and a cognitive economist. Your goal is to solve the
user's query with maximum accuracy and minimum computational cost (tokens).
You will follow a strict three-stage protocol.

Instruction: Analyze the following user query. Decompose it into a series of atomic,
sequential sub-questions that must be answered to solve the main query. Based on the
number and complexity of these sub-questions, provide a quantitative complexity
score from 1 (trivial) to 10 (extremely complex). Finally, propose a token budget
for the reasoning stage (Stage 3).

User Query:
"""
{{user_query}}
"""

Your Analysis Output (Format: JSON):
{
  "sub_questions": [
    "1. [First sub-question]",
    "2. [Second sub-question]",
    "..."
  ],
  "complexity_score": [Your score from 1 to 10],
  "reasoning_token_budget": [Proposed number of tokens, e.g., 50, 200, 500]
}


### STAGE 2: ECL Reduction (For RAG tasks only) ###
Instruction: You are now an information filtering expert. You have been provided
with the user query, the decomposed sub-questions from Stage 1, and a set of raw
retrieved documents. Your task is to read through the raw documents and extract
ONLY the precise sentences or facts that directly answer one or more of the
sub-questions. Discard all other information. Your output should be a concise,
clean set of facts.

User Query: {{user_query}}
Sub-questions: {{sub_questions_from_stage1}}
Raw Retrieved Documents:
"""
{{raw_documents}}
"""

Your Pruned Context Output:
[List of extracted, highly relevant facts and sentences]


### STAGE 3: GCL Application & Self-Correction ###
Instruction: You will now solve the user's query. Use the pruned context (if
available) and follow the plan of sub-questions from Stage 1. Generate a
step-by-step, Chain-of-Thought reasoning process. Adhere to the token budget.
After deriving the final answer, perform a final check to ensure your answer
is consistent with all reasoning steps and addresses all sub-questions.

User Query: {{user_query}}
Sub-questions: {{sub_questions_from_stage1}}
Pruned Context: {{pruned_context_from_stage2}}
Token Budget: {{token_budget_from_stage1}}

Your Final Response (Format: Markdown):
**Reasoning:**
Step 1: Answering '[First sub-question]'... [Your reasoning and intermediate answer for step 1]
Step 2: Answering '[Second sub-question]'... [Your reasoning and intermediate answer for step 2]
...

**Final Answer:**
[Your final, conclusive answer]

**Self-Correction Check:**
- All sub-questions addressed: [Yes/No]
- Final answer consistent with reasoning: [Yes/No]
\end{verbatim}

\section{CLAI-Tune Synthetic Dataset Samples}

Below are a few samples from the synthetic dataset used to fine-tune the CLAI-Tune model, demonstrating the ideal output formats for problems of varying complexity.

\subsection{Sample 1: Medium Complexity (GSM8K)}
\begin{lstlisting}[language=json, caption={Medium complexity sample from GSM8K.}, label={lst:sample_gsm8k}]
{
  "instruction": "Natalia sold 48 cupcakes in the morning. In the afternoon she sold half as many as she sold in the morning. In the evening she sold 15 cupcakes. How many cupcakes did she sell in total?",
  "output": "Here is the step-by-step solution:\n1. First, calculate the number of cupcakes sold in the afternoon. This is half of the morning sales, so 48 / 2 = 24 cupcakes.\n2. Next, calculate the total number of cupcakes sold. This is the sum of morning, afternoon, and evening sales: 48 + 24 + 15.\n3. 48 + 24 = 72.\n4. 72 + 15 = 87.\nSo, Natalia sold 87 cupcakes in total."
}
\end{lstlisting}

\subsection{Sample 2: High Complexity (MATH - Auto-decomposition)}
\begin{lstlisting}[language=json, caption={High complexity sample from MATH, demonstrating auto-decomposition.}, label={lst:sample_math}]
{
  "instruction": "Let $f(x) = x^3 - 6x^2 + 11x - 6$. Find the area of the region enclosed by the graph of $f(x)$, the x-axis, and the lines $x=1$ and $x=3$.",
  "output": {
    "analysis": "This problem requires finding the definite integral of a function, but first, I must determine if the function crosses the x-axis within the interval [1, 3], as this would require splitting the integral. I will decompose the problem into a plan.",
    "plan": [
      { "step": 1, "sub_problem": "Find the roots of the function $f(x) = x^3 - 6x^2 + 11x - 6 = 0$ to see where it crosses the x-axis." },
      { "step": 2, "sub_problem": "Check if any of the roots lie within the interval [1, 3]." },
      { "step": 3, "sub_problem": "Based on the roots, set up the correct definite integral(s) to calculate the total area. Remember that area must be positive, so use absolute values for parts of the function below the x-axis." },
      { "step": 4, "sub_problem": "Calculate the definite integral(s) to find the final area." }
    ]
  }
}
\end{lstlisting}

\subsection{Sample 3: RAG Task (LongBench QA)}
\begin{lstlisting}[language=json, caption={RAG task sample from LongBench QA.}, label={lst:sample_rag}]
{
  "instruction": "Based on the provided articles, what was the primary reason for the failure of the 'Great Leap Forward' campaign in China?",
  "input": "[A long list of retrieved documents, some relevant, some not...]",
  "output": "Based on the provided context, the primary reasons for the failure of the Great Leap Forward were:\n1. **Unrealistic agricultural targets:** The policy of 'taking steel as the key link' led to a massive diversion of agricultural labor to backyard steel furnaces, causing a sharp decline in grain production.\n2. **Lysenkoism:** The adoption of unscientific agricultural practices, such as close planting and deep plowing, promoted by Soviet biologist Trofim Lysenko, led to widespread crop failures.\n3. **Exaggerated reporting:** Local officials, under political pressure, vastly exaggerated grain yields, leading the central government to procure more grain than was actually available, leaving peasants with nothing."
}
\end{lstlisting}

\section{Experimental Hyperparameters}
To ensure the reproducibility of our experiments, we list the key hyperparameter settings.

\begin{itemize}
    \item \textbf{Models:}
    \begin{itemize}
        \item Teacher Model (for CLAI-Prompt \& Data Generation): \texttt{GPT-4o-2024-05-13}
        \item Student Model (for CLAI-Tune): \texttt{Meta-Llama-3-8B-Instruct}
        \item Baseline Models: \texttt{Meta-Llama-3-8B-Instruct}, \texttt{Meta-Llama-3-70B-Instruct}
    \end{itemize}
    \item \textbf{Fine-tuning (CLAI-Tune):}
    \begin{itemize}
        \item Framework: PyTorch FSDP
        \item Optimizer: AdamW
        \item Learning Rate: $2 \times 10^{-5}$
        \item Learning Rate Schedule: Cosine decay
        \item Batch Size: 128
        \item Epochs: 3
        \item Max Sequence Length: 4096
    \end{itemize}
    \item \textbf{Decoding Parameters:}
    \begin{itemize}
        \item All experiments used Greedy Decoding (temperature=0.0) to ensure consistency and comparability of results.
    \end{itemize}
    \item \textbf{Speculative Decoding Setup:}
    \begin{itemize}
        \item Draft Model: \texttt{TinyLlama-1.1B}
        \item Main Model: \texttt{Meta-Llama-3-8B-Instruct}
        \item Candidate Steps (K): 4
    \end{itemize}
\end{itemize}

\end{document}